\documentclass[conference]{IEEEtran}
\IEEEoverridecommandlockouts
\usepackage{cite}
\usepackage{amsmath,amsfonts}
\usepackage{subcaption}
\usepackage{graphicx}
\usepackage{textcomp}
\usepackage{xcolor}
\usepackage{url}
\usepackage{booktabs}
\usepackage{bbding}
\usepackage{color}
\usepackage{multirow}
\usepackage{enumitem}
\usepackage[accsupp]{axessibility}
\usepackage{makecell}
\usepackage{colortbl}
\definecolor{lightpink}{rgb}{1.0, 0.71, 0.76}
\definecolor{lightblue}{rgb}{0.68, 0.85, 0.9}
\usepackage{algpseudocode}

\usepackage{marvosym}
\usepackage[colorlinks,citecolor=blue,linkcolor=red,bookmarks=false]{hyperref}
\usepackage[capitalize]{cleveref}
\crefname{section}{Sec.}{Secs.}
\Crefname{section}{Section}{Sections}
\Crefname{table}{Table}{Tables}
\crefname{table}{Tab.}{Tabs.}

\newcommand{\boldstart}[1]{\vspace{0.1in}\noindent\textbf{#1}}

\def\BibTeX{{\rm B\kern-.05em{\sc i\kern-.025em b}\kern-.08em
    T\kern-.1667em\lower.7ex\hbox{E}\kern-.125emX}}
\begin{document}

\title{DriveGen3D: Boosting Feed-Forward Driving Scene Generation with Efficient Video Diffusion}

\author{
    Weijie Wang\textsuperscript{1,2}\textsuperscript{*} \quad
    Jiagang Zhu\textsuperscript{2,3}\textsuperscript{*\Letter} \quad
    Zeyu Zhang\textsuperscript{2} \quad
    Xiaofeng Wang\textsuperscript{2,4} \quad
    Zheng Zhu\textsuperscript{2}\textsuperscript{\Letter} \quad
    Guosheng Zhao\textsuperscript{2,4} \\
    Chaojun Ni\textsuperscript{2,5} \quad
    Haoxiao Wang\textsuperscript{1} \quad
    Guan Huang\textsuperscript{2} \quad
    Xinze Chen\textsuperscript{2} \quad
    Yukun Zhou\textsuperscript{2} \quad
    Wenkang Qin\textsuperscript{2} \\
    Duochao Shi\textsuperscript{1} \quad
    Haoyun Li\textsuperscript{2,4} \quad
    Yicheng Xiao\textsuperscript{4} \quad
    Donny Y. Chen\textsuperscript{6} \quad
    Jiwen Lu\textsuperscript{3} \\[0.3em]
    \textsuperscript{1}Zhejiang University \quad
    \textsuperscript{2}GigaAI \quad
    \textsuperscript{3}Tsinghua University \\
    \textsuperscript{4}Institute of Automation, Chinese Academy of Sciences \quad
    \textsuperscript{5}Peking University \quad
    \textsuperscript{6}Monash University \quad \\[0.3em]
    \textsuperscript{*}Equal contribution \quad \textsuperscript{\Letter}Corresponding authors \\[0.3em]
    Project Page: \href{https://lhmd.top/drivegen3d}{https://lhmd.top/drivegen3d}
    \vspace{-1.5em}
}

\maketitle

\begin{abstract}
We present DriveGen3D, a novel framework for generating high-quality and highly controllable dynamic 3D driving scenes that addresses critical limitations in existing methodologies. Current approaches to driving scene synthesis either suffer from prohibitive computational demands for extended temporal generation, focus exclusively on prolonged video synthesis without 3D representation, or restrict themselves to static single-scene reconstruction. Our work bridges this methodological gap by integrating accelerated long-term video generation with large-scale dynamic scene reconstruction through multimodal conditional control. DriveGen3D introduces a unified pipeline consisting of two specialized components: FastDrive-DiT, an efficient video diffusion transformer for high-resolution, temporally coherent video synthesis under text and Bird's-Eye-View (BEV) layout guidance; and FastRecon3D, a feed-forward module that rapidly builds 3D Gaussian representations across time, ensuring spatial-temporal consistency. DriveGen3D enable the generation of long driving videos (up to $800\times424$ at $12$ FPS) and corresponding 3D scenes, achieving state-of-the-art results while maintaining efficiency.
\end{abstract}


\section{Introduction}

The synthesis of 3D dynamic driving environments has emerged as a key research frontier in autonomous systems, driven by its wide-ranging applications in simulation, perception~\cite{wang2025transdiff}, and planning~\cite{zhao2026cov}. While recent advances in video generation~\cite{gao2024magicdrivedit,li2025psa,duan2026liveworld,wu2026light4d,zhang2025egolcd,zhang2025blockvid,luo2025univid,liu2025fpsattention,zhang2026panflow} and 3D scene reconstruction~\cite{streetgaussian, tian2024drivingforward,wang2025volsplat,liu2024rgbgrasp,wang2024adafsnet,liu2025trace} have made substantial progress, an integrated and efficient framework that unifies long-horizon video synthesis and large-scale 3D scene reconstruction under multi-modal control is still lacking.

Existing methodologies typically address either temporal coherence in video generation or spatial fidelity in scene reconstruction, but not both—often requiring high computational cost or suffering from limited scalability. For example, state-of-the-art diffusion-based models like MagicDriveDiT~\cite{gao2024magicdrivedit} can produce high-resolution driving sequences but require up to 30 minutes to generate a single 20s video, making them impractical for real-time use.

From the reconstruction perspective, optimization-based approaches~\cite{streetgaussian} are similarly time-consuming, often needing 30 minutes per scene. While recent feed-forward methods~\cite{charatan23pixelsplat, chen2024mvsplat, tian2024drivingforward, driv3r} have reduced reconstruction time to seconds, they remain limited in scale and rarely integrate with dynamic video generation pipelines.

To bridge this gap, we propose \textbf{DriveGen3D}, an efficient and unified framework that integrates two specialized modules: \textbf{FastDrive-DiT}, an accelerated video diffusion transformer for high-resolution driving video generation, and \textbf{FastRecon3D}, a feed-forward reconstruction pipeline that builds dynamic 3D scenes from multi-view video frames in real-time.

To address the latency in video generation, we analyze the diffusion process and identify that applying caching strategies~\cite{liu2024timestep} to the conditional branch maximizes speedup without compromising layout control. Furthermore, we addressed the computational bottleneck of the attention layers. By profiling the attention maps, we apply precise 8-bit quantization~\cite{zhang2025sageattention, zhang2024sageattention2} specifically to the cross-view attention modules, reducing inference time by over $2\times$ while preserving visual fidelity.

FastRecon3D enables rapid driving scene reconstruction. Unlike prior methods that treat frames independently~\cite{charatan23pixelsplat, chen2024mvsplat} or optimize each scene individually~\cite{streetgaussian}, we propose a recursive, feed-forward and temporal-aware approach. By employing cross-attention to fuse features from past and future contexts, our model reconstructs 3D primitives that are both spatially accurate and temporally coherent. This effectively mitigates motion blur in dynamic elements and ensures consistent geometry across the synthesized sequence.

We conduct extensive experiments on the nuScenes dataset \cite{Caesar_2020_CVPR_nuScenes} demonstrating that DriveGen3D outperforms state-of-the-art baselines in both video quality and 3D reconstruction accuracy. Specifically, Our framework achieves a PSNR of 23.25, an SSIM of 0.820, and an LPIPS of 0.176 on generated inputs, significantly outperforming previous state-of-the-art methods. Moreover, DriveGen3D reduces the end-to-end generation time to under 6 minutes, including video synthesis and 3D scene reconstruction, marking a substantial improvement in efficiency.
Our contributions can be summarized as follows:

\begin{figure*}
\centering
\includegraphics[width=1.0\linewidth]{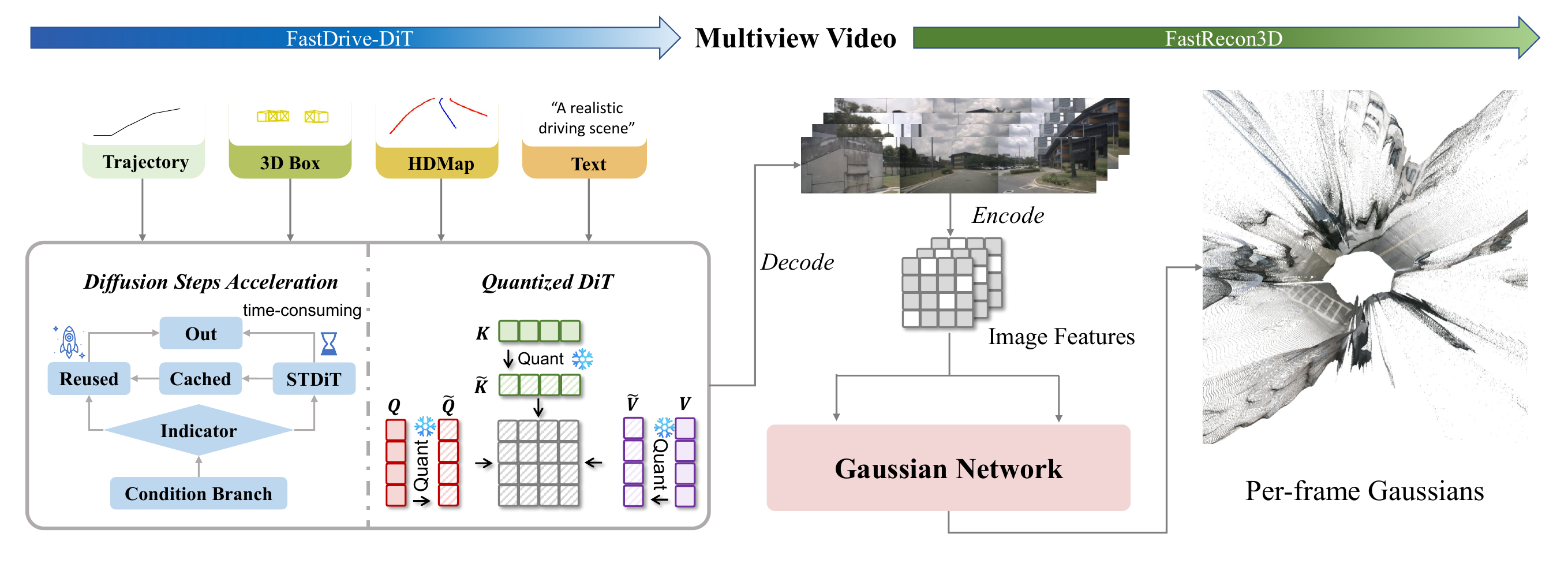}
\caption{\label{fig:method}\textbf{Overview of DriveGen3D.} (a) Given textual and BEV layout conditions, our model first employs an accelerated Video Diffusion Transformer to synthesize a long driving video. (b) Next, a per-frame 3D Gaussian Splatting representation is utilized to construct entire scene from the generated video frames.}
\vspace{-1.5em}
\end{figure*}

\begin{itemize}[itemsep=0pt, parsep=0pt]
    \item We introduce DriveGen3D, a unified framework for efficient 3D driving scene generation that combines video synthesis and feed-forward 3D reconstruction under multimodal control via text and BEV layout.

    \item We propose FastDrive-DiT, a high-performance video diffusion transformer equipped with step caching and quantized attention, achieving over $2\times$ inference speedup while maintaining visual fidelity. We also develop FastRecon3D, a feed-forward 3D scene reconstruction module based on temporal-aware Gaussian splatting, enabling fast and scalable scene modeling from generated videos.

    \item DriveGen3D achieves high-quality video generation and complete 3D reconstruction for large-scale street scenes within 6 minutes, significantly outperforming the state-of-the-art methods in terms of photorealism, scalability and efficiency.

\end{itemize}

\section{Related Work}

\boldstart{Video generation for driving scene}
Recent advancements in street view generation and autonomous driving scene synthesis have significantly improved the fidelity and controllability of synthetic data. MagicDrive \cite{gao2023magicdrive} introduces a framework for street view generation with diverse 3D geometry controls, such as camera poses and 3D bounding boxes, enhancing tasks like BEV segmentation and 3D object detection through cross-view consistency. MagicDriveDiT \cite{gao2024magicdrivedit} extends this by addressing high-resolution, long video generation for autonomous driving, leveraging flow matching and spatial-temporal conditional encoding to achieve superior scalability and control. 
UniScene \cite{li2024uniscene} unifies the generation of semantic occupancy, video, and LiDAR data, employing a progressive generation process to reduce complexity and improve downstream task performance.Together, these works advance the field of autonomous driving by providing scalable, controllable, and high-fidelity synthetic data generation frameworks.

\boldstart{Reconstruction for driving scene}
The reconstruction of dynamic driving scenes~\cite{streetgaussian} has emerged as a critical task in autonomous systems and immersive environment modeling. Contemporary approaches predominantly leverage Gaussian splatting-based representations due to their inherent balance between rendering efficiency and geometric expressiveness. Early methodologies in this domain adopted optimization-based paradigms, exemplified by works such as StreetGaussian~\cite{streetgaussian}. These frameworks optimize scene representations per instance for specific street segments (typically under 100 meters in scale) through iterative refinement processes spanning approximately 30 minutes. 
Recent advancements have shifted toward feed-forward architectures~\cite{charatan23pixelsplat,chen2024mvsplat,wang2025zpressor, shi2025pmloss, wang2025volsplat, ni2025wonderturbo, xu2025resplat} to enable rapid 3D reconstruction. Methods like PixelSplat~\cite{charatan23pixelsplat} and MVSplat~\cite{chen2024mvsplat} employ large pretrained networks to directly infer Gaussian parameters from multi-view inputs, reducing reconstruction time from minutes to seconds. Though these approaches demonstrate generalizability across scenes, they often sacrifice reconstruction fidelity in geometrically complex regions or under sparse observational constraints. Parallel innovations address the temporal dimension of driving scenes: InfiniCube~\cite{lu2024infinicube} extends the Scube~\cite{ren2024scube} framework to 3D street generation, disentangling dynamic vehicles from static backgrounds via hybrid voxel-video control mechanisms. Drive3R~\cite{driv3r} adapts the architecture of Spann3R~\cite{wang2024spann3r} for per-frame 3D scene reconstruction through temporal consistency priors. DrivingForward~\cite{tian2024drivingforward} enhances sparse-view reconstruction robustness by jointly learning pose estimation and depth prediction modules within its network architecture. These advancements collectively highlight two persistent limitations: 

\begin{itemize}[itemsep=0pt, parsep=0pt]
\item Existing feed-forward 3D reconstruction methods operate at constrained spatial resolutions (typically $\leq 512\times512$) 

\item Prior works on generative models for conditional scene synthesis required substantial computational resources and complex framework integration, which hindered their widespread adoption.
\end{itemize}

\section{Method}

\subsection{Overview}

DriveGen3D is an integrated 3D driving scene generation framework composed of two key components as illustrated in Figure~\ref{fig:method}. The pipeline begins with FastDrive-DiT, which synthesizes high-resolution, temporally coherent driving videos under conditional guidance. These generated videos are then passed to FastRecon3D, which reconstructs dynamic 3D scenes in a feed-forward manner using temporal-aware Gaussian splatting. Together, these modules enable rapid and scalable 3D scene generation suitable for real-time simulation and autonomous driving applications.

The architecture of our efficient video diffusion module, dubbed FastDrive-DiT, integrates two targeted acceleration strategies to address the high inference cost of driving video generation, leading to a speedup of more than $2\times$.
FastRecon3D proposes a temporally context-visible fusion module based on the static feed-forward framework, resulting in better temporally consistent per-frame Gaussians.

\subsection{FastDrive-DiT}

Generating 3D driving scenes, especially in the autonomous driving domain, is notoriously time-consuming due to the multi-view nature of the data. The video generation step is particularly costly because of the underlying diffusion process. For instance, MagicDriveDiT~\cite{gao2024magicdrivedit} can take up to 30 minutes to produce a video of driving scene. To address this inefficiency, we propose FastDrive-DiT, an enhanced and lightweight video diffusion model built on MagicDriveDiT with targeted acceleration strategies.

\boldstart{Diffusion steps acceraleration}. 
To accelerate the generation of long and high-resolution videos in MagicDriveDiT, we integrate TeaCache\cite{liu2024timestep}, a training-free caching approach that enhances the inference speed of diffusion models. TeaCache leverages timestep embeddings to modulate noisy inputs, approximating the fluctuating differences in model outputs across timesteps. By introducing a rescaling strategy, it refines these differences, enabling efficient output caching with minimal computational overhead. Experiments show that TeaCache achieves up to $4.41\times$ acceleration while maintaining visual quality, making it an effective solution for balancing speed and performance in video generation.

In the original TeaCache\cite{liu2024timestep} implementation, coefficients are computed from both the conditional and unconditional branches of the diffusion model. However, in our adaptation for MagicDriveDiT, we calculate coefficients exclusively from the conditional branch. This optimization not only reduces the forwarding time of TeaCache but also ensures no obvious degradation in performance, further enhancing the efficiency of the video generation process.

\boldstart{Quantized DiT}. Considering the high cost of the attention part in the video DiT, we use SageAttention\cite{zhang2025sageattention,zhang2024sageattention2} for quantization during the inference procedure. 
By focusing on quantizing attention, SageAttention\cite{zhang2025sageattention} achieves about $2.1\times$ higher OPS (operations per second) compared to FlashAttention2 \cite{dao2023flashattention2}, while maintaining accuracy across diverse models. SageAttention2\cite{zhang2024sageattention2} further enhances efficiency by introducing INT4 quantization for $Q$ and $K$ matrices, FP8 for $P$ and $V$, and precision-enhancing techniques like outlier smoothing and FP32 Matmul buffers. It achieves 3$\times$ higher OPS than FlashAttention2~\cite{dao2023flashattention2}, with negligible accuracy loss.

We carefully analyze the inference profiling and visualize $Q\times K \times V$ of different attention blocks, pointing out potential improvement techniques. In this paper, applying sageattention to the transformer blocks saves an additional 30 seconds of inference time with nearly no performance degradation. Experiment details are in section \ref{Quantization_with_SageAttention_exp}.

\begin{figure}
\centering
\includegraphics[width=1.0\linewidth]{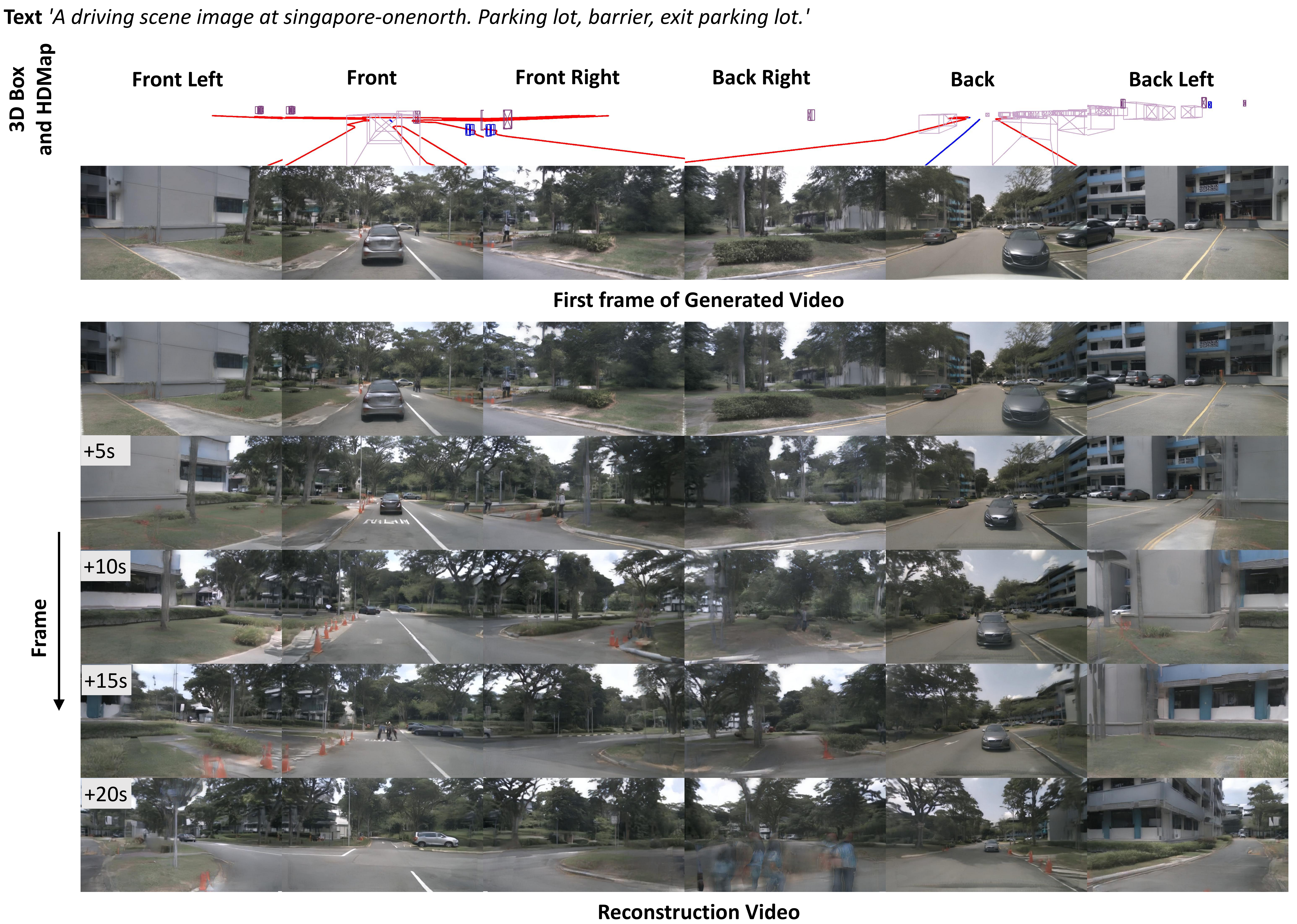}
\caption{Visualization of the multi-view reconstructed video from a generated 3D scene.}
\label{fig:generation_reconstruction}
\vspace{-1.5em}
\end{figure}
\subsection{FastRecon3D}

While the aforementioned methods enable realistic driving scene generation, applications like simulation require complete 3D scene models. To enable rapid novel scene synthesis, we introduce FastRecon3D, a feed-forward reconstruction module that avoids costly optimization while preserving quality. Built upon temporal-aware Gaussian Splatting, FastRecon3D reconstructs the 3D geometry of each frame 3D geometry by leveraging both past and future contexts, enabling fast and consistent 3D scene generation.

\boldstart{Per-Frame 3D Representation}. Existing methods like Street Gaussians~\cite{streetgaussian} and Infinicube~\cite{lu2024infinicube} attempt dynamic scene reconstruction through object-scene separation but suffer from two critical limitations: 1) Dynamic elements in static backgrounds (e.g., pedestrians, cyclists) suffer motion blur artifacts 2) Computational constraints prevent large-scale scene reconstruction. Following DrivingForward~\cite{tian2024drivingforward}, we propose a temporal-aware Gaussian Splatting formulation that reconstructs per-frame 3D Gaussian models while maintaining 3D consistency.
In our approach, each time step is represented by a set of 3D Gaussian primitives:

\begin{equation}
    G_i^t=\{\mathcal{G}_i^t\}_{i=1}^{N_t}=\{(\mathbf{\mu}_{i}, \mathbf{\Sigma}_{i}, \mathbf{\alpha}_{i}, \mathbf{c}_{i})\}_{i=1}^{N_t}.
\end{equation}

\boldstart{Recursively reconstruction from videos}.  We employ the decoder of DrivingForward~\cite{tian2024drivingforward} as the Gaussian Network to estimate Gaussian parameters, and use a pre-trained image encoder~\cite{oquab2023dinov2} to extract features.

After the frame generation of the $t+\Delta$ timestep, we extract all frames from time steps $t-\Delta$ to $t+\Delta$ to reconstruct the scene at time $t$. Our model first extracts features $\{F_i^t\}_{i=1}^{N_t}$ from each frame. For each timestep $t$, given multi-view features $\{F_i^t\}_{i=1}^{N_t}$ and temporal neighbors $\{F_i^{t\pm\Delta}\}_{i=1}^{N_t}$, we use cross-attention to fuse information from other frames into $\{F_i^t\}_{i=1}^{N_t}$:

\begin{equation}
    \{ {F'}_i^t \}_{i=1}^{N_t} = \text{Attention}(Q, K, V),
\end{equation}

\begin{equation}
   Q \leftarrow \{F_i^t\}_{i=1}^{N_t}, \quad K, V \leftarrow \{F_i^{t\pm\Delta}\}_{i=1}^{N_t}.
\end{equation}

\begin{table}[t]
\centering
\caption{
Comparison of our method against prior feed-forward and optimization-based methods. The last two rows show novel view rendering performance with either GT or generated video input.
}
\resizebox{\linewidth}{!}{
\begin{tabular}{l|l|ccc}
\toprule
Type & Method & PSNR $\uparrow$ & SSIM $\uparrow$ & LPIPS $\downarrow$ \\
\midrule
\multirow{2}{*}{Static} & MVSplat~\cite{chen2024mvsplat}         & 22.83 & 0.629 & 0.317 \\
                        & pixelSplat~\cite{charatan23pixelsplat}      & 25.00 & 0.727 & 0.298 \\
\midrule
\multirow{8}{*}{Dynamic}& UniPad~\cite{yang2024unipad}         & 16.45 & 0.375 & -     \\
                        & SelfOcc~\cite{huang2024selfocc}      & 18.22 & 0.464 & -     \\
                        & EmerNeRF~\cite{emernerf}      & 20.95 & 0.585 & -     \\
                        & DistillNeRF~\cite{wang2024distillnerf}    & 20.78 & 0.590 & -     \\
                        & DrivingForward~\cite{tian2024drivingforward}  & 21.69 & 0.742 & 0.226 \\
\cmidrule(l){2-5}
                        & Ours (w/ GT images) & \textbf{23.87} & 0.814 & \textbf{0.169} \\
                        & Ours (w/ GEN images)  & 23.25 & \textbf{0.820} & 0.176 \\
\bottomrule
\end{tabular}
}
\label{tab:combined_recon}
\vspace{-1.5em}
\end{table}

Then the network predicts Gaussian parameters and save it as a 3D model:
\begin{equation}
\mathcal{G}^t = \{ \mu^t, \Sigma^t, \alpha^t, c^t \} = \Psi\Bigl( \{{F'}_i^{t-\Delta},\, {F'}_i^t,\, {F'}_i^{t+\Delta}\}_{i=1}^{N_t} \Bigr).
\end{equation}

By leveraging both past and future context, this recursive reconstruction method effectively captures dynamic scene elements while maintaining high spatial fidelity. As a result, our approach produces complete 3D models in a matter of seconds, meeting the demanding requirements of simulation and other real-time applications without compromising on quality.

\section{Experiments}

\subsection{Experimental Setup}

\boldstart{Dataset}. 
The training dataset is obtained from the nuScenes dataset \cite{Caesar_2020_CVPR_nuScenes}. It consists of 700 training videos and 150 validation videos. For the 3D scene reconstruction model, we split the dataset into 20,000 short sequences for training.

\begin{figure}
\centering
\includegraphics[width=\linewidth]{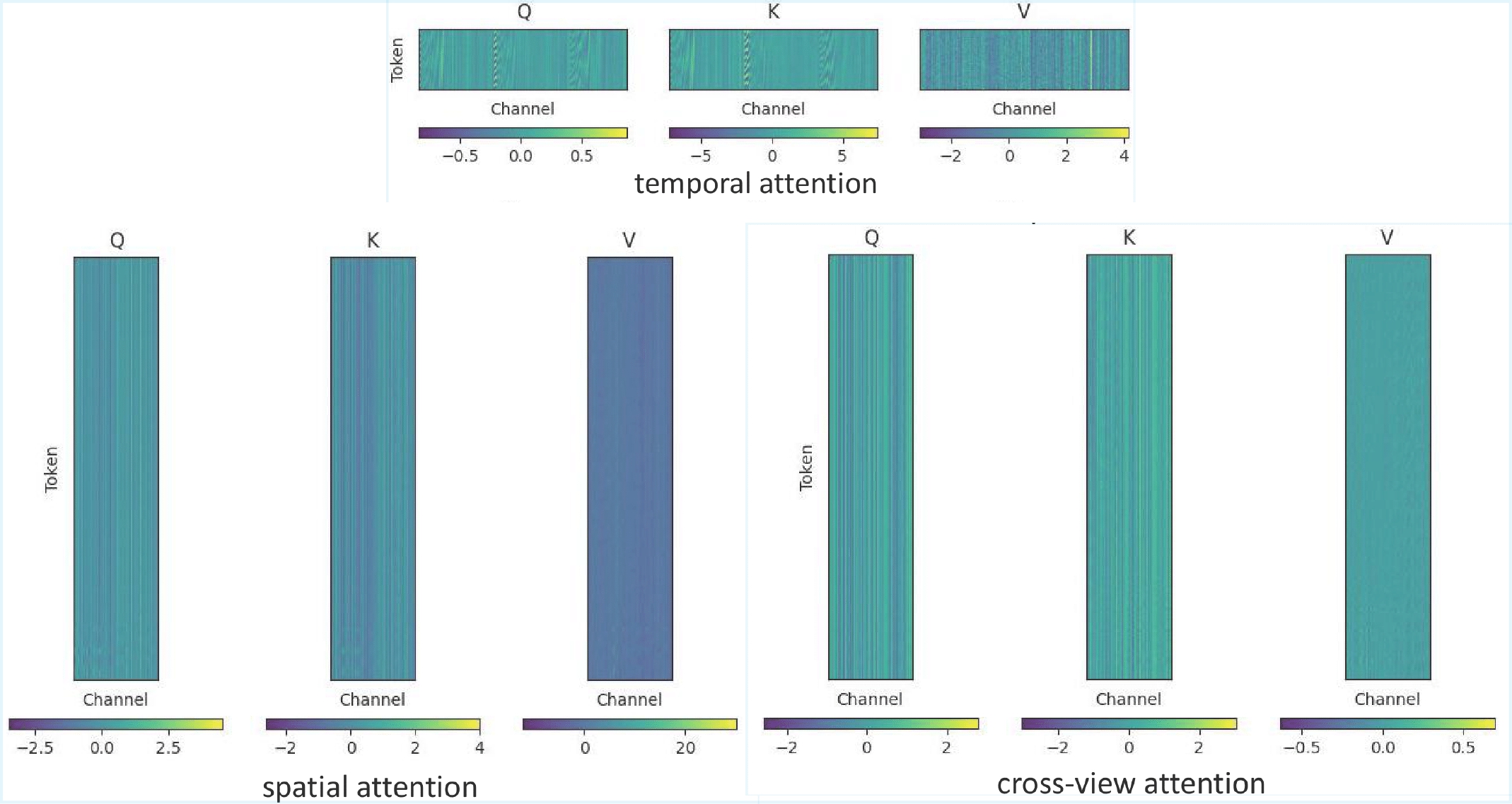}
\caption{\label{fig:qkv} Typical examples of the data distribution of tensors in different attention blocks of MagicDriveDiT.}
\vspace{-1em}
\end{figure}

\begin{figure}[t]
    \centering
    \begin{subfigure}[b]{0.32\linewidth} %
        \centering
        \includegraphics[width=\linewidth]{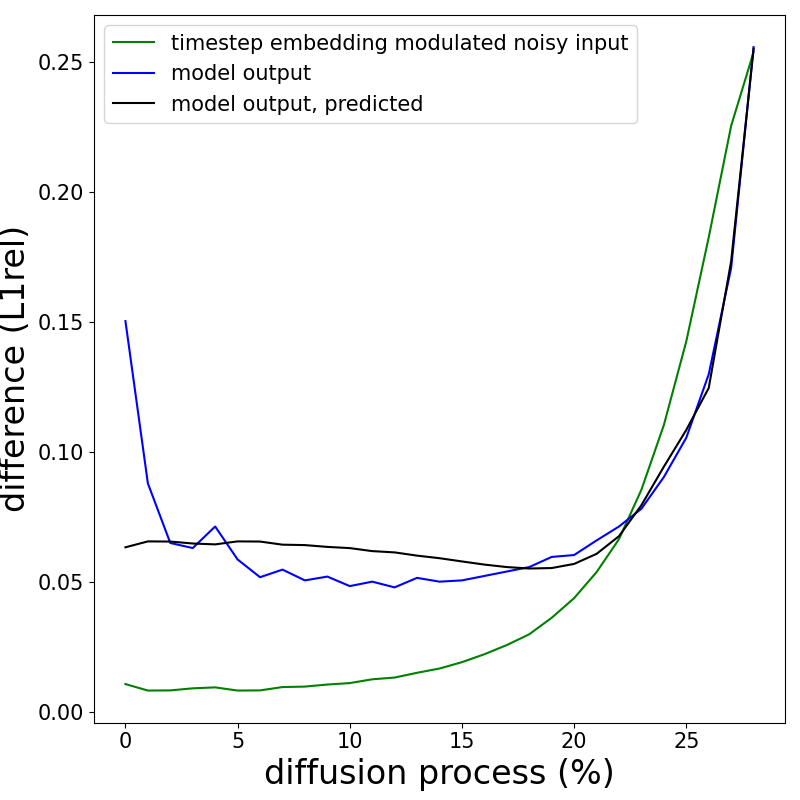}
        \caption{All}
        \label{fig:teacache-all}
    \end{subfigure}%
    \hfill
    \begin{subfigure}[b]{0.32\linewidth}
        \centering
        \includegraphics[width=\linewidth]{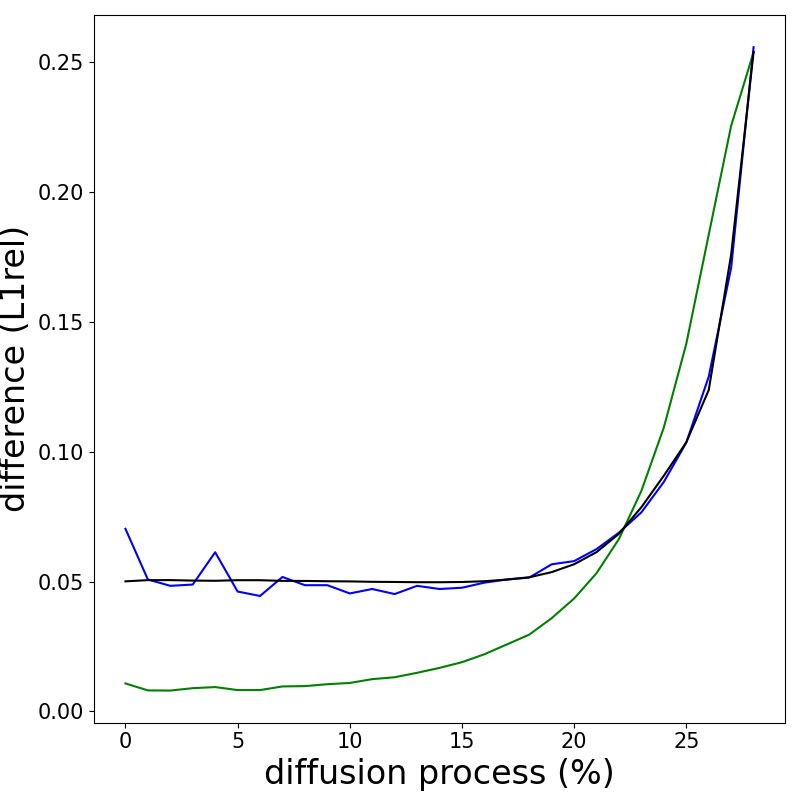}
        \caption{Condition}
        \label{fig:teacache-condition}
    \end{subfigure}%
    \hfill
    \begin{subfigure}[b]{0.32\linewidth}
        \centering
        \includegraphics[width=\linewidth]{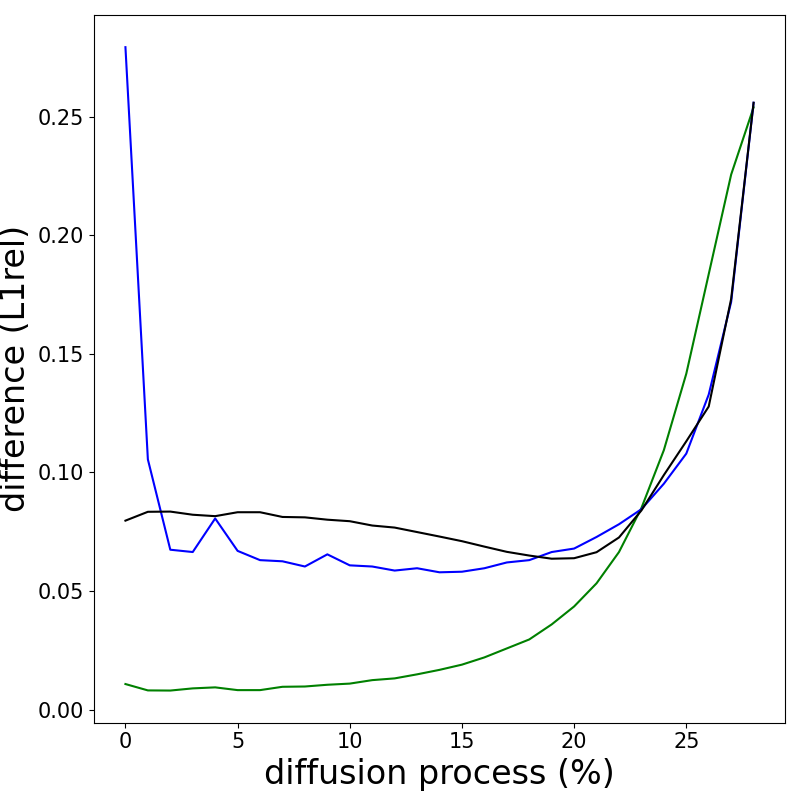}
        \caption{Uncondition}
        \label{fig:teacache-uncondition}
    \end{subfigure}
    
    \caption{Visualization of input differences and output differences in consecutive timesteps. We plot \textit{all}, \textit{condition} and \textit{uncondition} separately.}
    \label{fig:main_fig_teacache}
\vspace{-1.5em}
\end{figure}

\boldstart{Evaluation Metrics}. 
For video-generation stage, we evaluate both the realism and controllability in street-view video generation. For video generation, we adhere to the benchmarks from \cite{W-CODA2024}. To measure video quality, we use the Frechet Video Distance (FVD). Regarding controllability, we employ mAP from 3D object detection and mIoU from BEV segmentation. For both of these tasks, we utilize BEVFormer~\cite{li2022bevformer}, which is a video-based perception model. We generate corresponding videos for the annotations in the validation set. Then, we apply the aforementioned metrics and use perception models pre-trained on real-world data for evaluation.

For 3D scene reconstruction stage, we adopt the view synthesis task to assess reconstruction quality, following the evaluation protocol established in Driving Forward \cite{tian2024drivingforward}. Given sequential video frames from timestamps $t-\Delta$ and $t+\Delta$ (In our implementation, $\Delta=2$) as input, our model synthesizes the intermediate frame at timestamp $t$ for quantitative comparison against ground truth. We report Peak Signal-to-Noise Ratio (PSNR), Structural Similarity Index Measure (SSIM), and Learned Perceptual Image Patch Similarity (LPIPS) in Table \ref{tab:combined_recon}. 
To benchmark performance, we compare our DriveGen3D reconstruction model against the Driving Forward baseline (evaluate on ground truth images) across same validation frames and synthesized video outputs. This dual evaluation strategy disentangles reconstruction capability from generation artifacts, providing comprehensive insights into geometric and appearance recovery accuracy.

\boldstart{Implementation Details}. We train the model with a resolution of $800\times424$. Inference is performed on NVIDIA H20 GPUs. When assessing the time cost of our proposed method, the baselines for the video generation model are MagicDriveDiT~\cite{gao2024magicdrivedit}(17f) and MagicDriveDiT~\cite{gao2024magicdrivedit}(233f). The 3D scene reconstruction stage is trained on a single NVIDIA H20 GPU for 2 days.

\begin{table}
\centering
\caption{Time cost of different attention blocks of MagicDriveDiT during inference.}
\resizebox{1\linewidth}{!}{
\begin{tabular}{c|ccccc}
\toprule
Total  & spatial & temporal & cross-view & cross & other \\
\midrule
615 s & 104 s  & 82 s & 163 s & 62 s & 204 s\\
\bottomrule
\end{tabular}}
\label{tab:time_profiling}
\vspace{-1em}
\end{table}
\begin{table}
\centering
\caption{Acceralerating the inference process of MagicDriveDiT.  17f and 233f denote the frames count of generated videos.
}
\resizebox{1\linewidth}{!}{
\begin{tabular}{l|cccc}
\toprule
& FVD $\downarrow$ & mAP $\uparrow$ & mIoU$\uparrow$ & Time cost  17f/233f $\downarrow$
\\ \midrule
MagicDriveDiT  & 111.58 & 17.10 & 21.92 & 211s/615 s \\
Ours (w/o Quant)  & 125.70 & 16.60 & 21.27  & 64 s/309 s \\
Ours  &125.88 & 16.72 & 21.24   & \textbf{58 s/278 s} \\
\bottomrule
\end{tabular}}
\label{tab:Acceralerating_teacache_sageattention}
\vspace{-1.5em}
\end{table}

\subsection{Experimental Results}

\subsubsection{Novel View Synthesis}
Table~\ref{tab:combined_recon} provides a comprehensive comparison of DriveGen3D against both optimization-based dynamic methods and feed-forward reconstruction methods. Notably, when using generated images instead of ground truth, DriveGen3D maintains competitive reconstruction quality with a PSNR of 22.84 and achieves the highest SSIM of 0.811, demonstrating strong temporal coherence and structure preservation in generated scenes. This suggests that despite operating on synthetic inputs, DriveGen3D produces reliable and structurally consistent 3D reconstructions, validating the effectiveness of its end-to-end pipeline.

\subsubsection{Experimental Analysis and Ablation Study}
We construct experiments to demonstrate the effectiveness of FastDrive-DiT.

\boldstart{Diffusion steps acceraleration}. 
\label{TeaCache_acceraleration_exp}
Firstly, we visualize the input differences and output differences in consecutive timesteps in Figure \ref{fig:main_fig_teacache}. It is observed that
the default configurations of TeaCache~\cite{liu2024timestep}, i.e. \textit{all}, exhibits a U-shape for model output, with downward tend initially, nearly constant for the middle and upward until the end. The same is for the \textit{uncondition} branch. The \textit{condition} branch exhibits a slightly different phenomenon, with the start of model output much smaller. We also apply simple polynomial fitting to fit a relationship between model input and output and the use these cofficients to predict model output according to the input. 
\begin{figure}
\centering
\includegraphics[width=\linewidth]{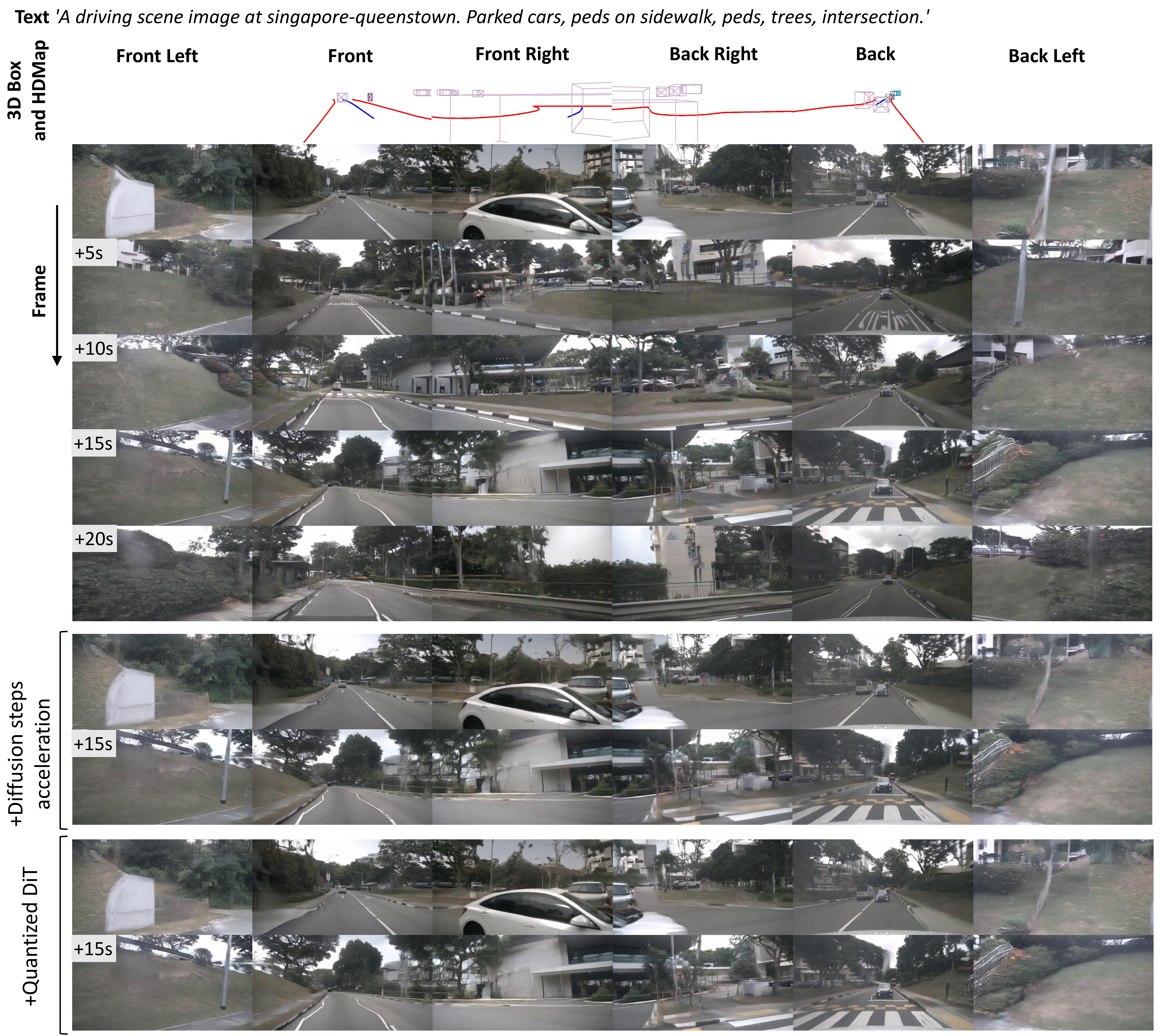}
\caption{Comparison of video generation for MagicDriveDiT, Diffusion steps acceleration and Quantized DiT.}
\label{fig:vis_base_teacache_sage}
\vspace{-1.5em}
\end{figure}
\begin{figure}
    \centering
    \includegraphics[width=\linewidth]{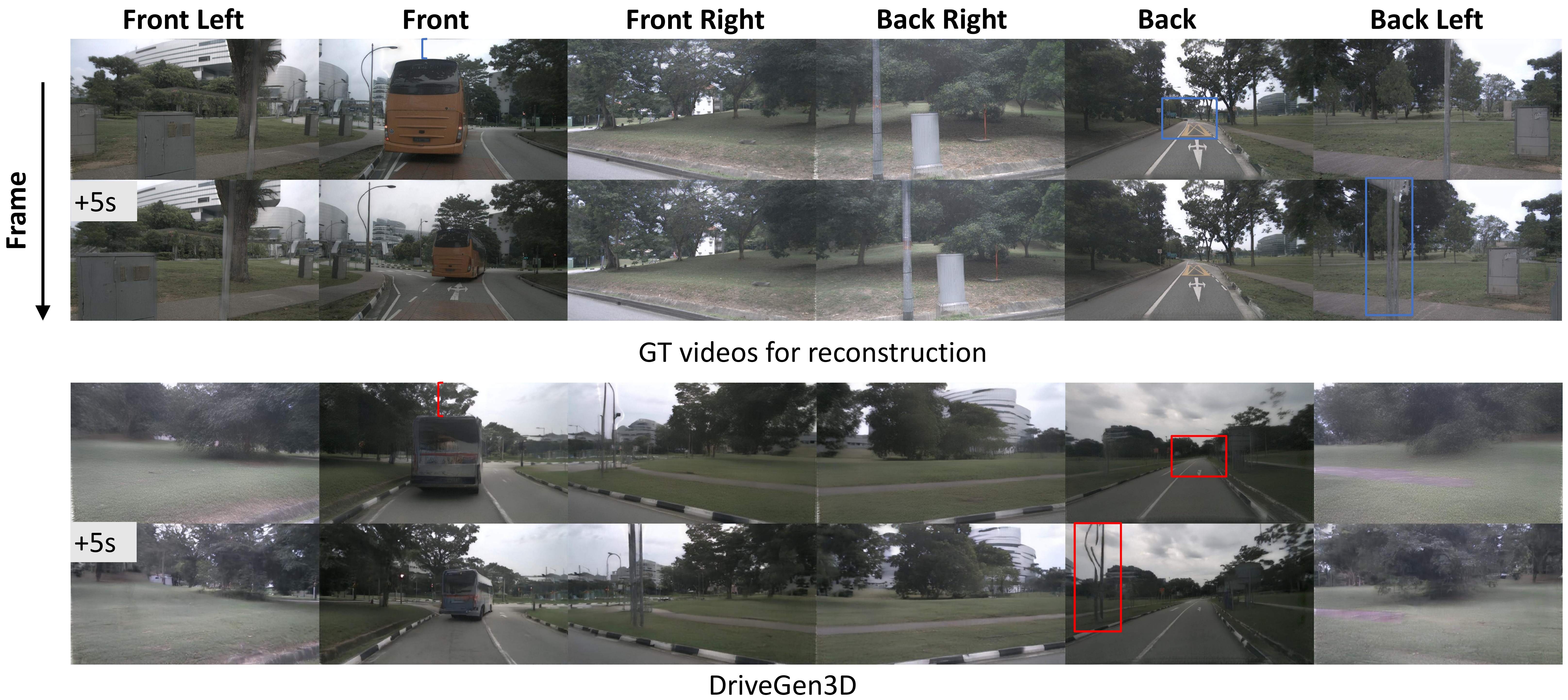}
    \caption{Comparison of GT videos and generated video for reconstruction.}
    \label{fig:cmp_gt_generation_for_reconstruction}
\vspace{-1.75em}
\end{figure}
As shown in the black lines in Figure \ref{fig:main_fig_teacache}, the quality of fitting is best for the \textit{condition} branch, while \textit{all} and \textit{uncondition} do not fit well for the start. We attribute this to the different input and output relationships in the start of diffusion process. So we finally only apply TeaCache to the \textit{condition} branch of MagicDriveDiT. As shown in Table \ref{tab:Acceralerating_teacache_sageattention}, equiping MagicDriveDiT with TeaCache for \textit{condition} branch has a speedup of nearly $3\times$ and $2\times$ for 17 frames and 233 frames. The perception metrics, mAP and mIoU only shows slight decrease.

\boldstart{Quantized DiT}. \label{Quantization_with_SageAttention_exp} Secondly, we profile the time cost of MagicDriveDiT. As shown in Table ~\ref{tab:time_profiling}, cross-view attention is identified as the most computationally costly process in MagicDriveDiT. We further analyze different attention components in MagicDriveDiT and plot the distribution of $Q, K, V$, as depicted in Figure \ref{fig:qkv}. Notably, $Q$ and $K$ exhibit a trend similar to that in Figure 4 of SageAttention2~\cite{zhang2024sageattention2}. An interesting phenomenon is observed: the numeric range of $V$ follows the order: spatial $>$ temporal $>$ cross-view. Specifically, the cross-view range is 10 times smaller than the spatial range and 5 times smaller than the temporal range. Given the principle of quantization, a smaller range is more conducive to quantization. Therefore, a promising solution is not only to apply SageAttention~\cite{zhang2025sageattention} to cross-view attention but also substitute the high-precision quantization method for V with a lower-resolution one. Concretely, for $V$, the FP8, E4M3 data type can be replaced with FP8, E5M2 data type and INT8. We leave the latter for future attempt. In this paper, applying sageattention to the transformer blocks saves an additional 30 seconds (233 frames) of inference time after diffusion steps acceraleration while maintaining performance.

\subsubsection{Visualization and Comparison}

We provide visual results of DriveGen3D alongside ablation experiments.

\boldstart{Visualization}. In Figure \ref{fig:vis_base_teacache_sage}, we compare the generated videos of baseline model, baseline model with diffusion steps acceraleration and further quantization with SageAttention. With the proposed techniques, no obvious change is observed. It can still generate video longer than 15 s with a much faster speed. This highlights the efficiency of \textit{DriveGen3D}.

\boldstart{Full pipeline results}. In Figure \ref{fig:generation_reconstruction}, we show a typical output result of our framework. \textit{DriveGen3D} can generate and reconstruct street scenes for more than 20s, 12FPS.

\boldstart{Comparison with raw videos}. By default, our reconstruction stage takes the generated videos as input. We compare with DrivingForward to validate the difference of raw videos and generated videos. As evidenced in Table \ref{tab:combined_recon}, DriveGen3D maintains fidelity in video generation that is essentially comparable to existing methodologies with GT images, exhibiting only a marginal degradation. Qualitative analysis is shown in Figure \ref{fig:cmp_gt_generation_for_reconstruction}. In the front view image of reconstruction videos, the spatial location of generated bus is different from gt videos, we check the 3D Box layout and find the generation videos does not match well. In the view of back left, the generation images is blurry, underscoring the needs of optimizing video generation model for reconstruction. 
Moreover, both GT videos and generation videos display drawbacks of flickering pole, pointing out the flaw of second stage reconstruction model.

\section{Conclusion}
We present DriveGen3D, an efficient framework for synthesizing long-duration driving videos from text prompts and BEV layouts, and generating high-quality large-scale dynamic scenes. Our approach marks a significant advancement in world modeling by bypassing conventional voxel-based generation paradigms~\cite{lu2024infinicube} through a novel integration of longitudinal video generation and scene reconstruction modules. This architecture enables faithful reproduction of real-world driving scenarios and thus paves the way for novel applications in autonomous vehicle simulation and dynamic world modeling.

\bibliographystyle{IEEEbib}
\bibliography{reference}

@String(CVPR = {IEEE Conf. Comput. Vis. Pattern Recog.})

@String(ECCV = {Eur. Conf. Comput. Vis.})

@String(NIPS = {Adv. Neural Inform. Process. Syst.})

@String(ICLR = {Int. Conf. Learn. Represent.})

@String(AAAI = {AAAI Conf. Artif. Intell.})

@String(NEURIPS = {Adv. Neural Inform. Process. Syst.})

@String(CVPR  = {CVPR})

@String(ECCV  = {ECCV})

@String(NIPS  = {NeurIPS})

@String(ICLR  = {ICLR})

@String(AAAI = {AAAI})

@article{gao2024magicdrivedit,
  title={MagicDriveDiT: High-Resolution Long Video Generation for Autonomous Driving with Adaptive Control},
  author={Gao, Ruiyuan and others},
  journal={arXiv preprint arXiv:2411.13807},
  year={2024}
}

@inproceedings{streetgaussian,
  title={Street gaussians: Modeling dynamic urban scenes with gaussian splatting},
  author={Yan, Yunzhi and Lin, Haotong and Zhou, Chenxu and Wang, Weijie and Sun, Haiyang and Zhan, Kun and Lang, Xianpeng and Zhou, Xiaowei and Peng, Sida},
  booktitle=ECCV,
  pages={156--173},
  year={2024},
  organization={Springer}
}

@inproceedings{wang2025transdiff,
  title={Transdiff: Diffusion-based method for manipulating transparent objects using a single rgb-d image},
  author={Wang, Haoxiao and Zhou, Kaichen and Gu, Binrui and Feng, Zhiyuan and Wang, Weijie and Sun, Peilin and Xiao, Yicheng and Zhang, Jianhua and Dong, Hao},
  booktitle={ICRA},
  year={2025}
}

@article{zhao2026cov,
  title={CoV: Chain-of-View Prompting for Spatial Reasoning},
  author={Zhao, Haoyu and others},
  journal={arXiv preprint arXiv:2601.05172},
  year={2026}
}

@article{li2025psa,
  title={PSA: Pyramid Sparse Attention for Efficient Video Understanding and Generation},
  author={Li, Xiaolong and Gu, Youping and Lin, Xi and Wang, Weijie and Zhuang, Bohan},
  journal={arXiv preprint arXiv:2512.04025},
  year={2025}
}

@article{emernerf,
  title={Emernerf: Emergent spatial-temporal scene decomposition via self-supervision},
  author={Yang, Jiawei and others},
  journal={arXiv preprint arXiv:2311.02077},
  year={2023}
}

@article{liu2024rgbgrasp,
  title={RGBGrasp: Image-based Object Grasping by Capturing Multiple Views During Robot Arm Movement with Neural Radiance Fields},
  author={Liu, Chang and others},
  journal={IEEE Robotics and Automation Letters},
  year={2024}
}

@inproceedings{wang2024adafsnet,
  title={Adafsnet: Time series classification based on convolutional network with a adaptive and effective kernel size configuration},
  author={Wang, Haoxiao and others},
  booktitle={IJCNN},
  year={2024}
}

@article{gao2023magicdrive,
  title={Magicdrive: Street view generation with diverse 3d geometry control},
  author={Gao, Ruiyuan and others},
  journal={arXiv preprint arXiv:2310.02601},
  year={2023}
}

@article{lu2024infinicube,
  title={InfiniCube: Unbounded and Controllable Dynamic 3D Driving Scene Generation with World-Guided Video Models},
  author={Lu, Yifan and others},
  journal={arXiv preprint arXiv:2412.03934},
  year={2024}
}

@article{tian2024drivingforward,
  title={Drivingforward: Feed-forward 3d gaussian splatting for driving scene reconstruction from flexible surround-view input},
  author={Tian, Qijian and others},
  journal={arXiv preprint arXiv:2409.12753},
  year={2024}
}

@inproceedings{charatan23pixelsplat,
      title={pixelSplat: 3D Gaussian Splats from Image Pairs for Scalable Generalizable 3D Reconstruction},
      author={David Charatan and others},
      year={2024},
      booktitle=CVPR,
}

@article{chen2024mvsplat,
    title   = {MVSplat: Efficient 3D Gaussian Splatting from Sparse Multi-View Images},
    author  = {Chen, Yuedong and others},
    journal = {arXiv preprint arXiv:2403.14627},
    year    = {2024},
}

@inproceedings{
      ren2024scube,
      title={SCube: Instant Large-Scale Scene Reconstruction using VoxSplats},
      author={Ren, Xuanchi and others},
      booktitle=NEURIPS,
      year={2024},
}

@article{driv3r,
  title={Driv3R: Learning Dense 4D Reconstruction for Autonomous Driving}, 
  author={Fei, Xin and others},
  journal={arXiv preprint arXiv:2412.06777},
  year={2024}
}

@article{wang2024spann3r,
    title={3D Reconstruction with Spatial Memory},
    author={Wang, Hengyi and Agapito, Lourdes},
    journal={arXiv preprint arXiv:2408.16061},
    year={2024}
}

@InProceedings{Caesar_2020_CVPR_nuScenes,
    author = {Caesar, Holger and others},
    title = {nuScenes: A Multimodal Dataset for Autonomous Driving},
    booktitle = CVPR,
    month = {June},
    year = {2020}
}

@article{liu2025trace,
  title={Trace anything: Representing any video in 4d via trajectory fields},
  author={Liu, Xinhang and Xiao, Yuxi and Chen, Donny Y and Feng, Jiashi and Tai, Yu-Wing and Tang, Chi-Keung and Kang, Bingyi},
  journal={arXiv preprint arXiv:2510.13802},
  year={2025}
}

@inproceedings{W-CODA2024,
  title={Challenge report: Track 2 of multimodal perception and comprehension of corner cases in autonomous driving},
  author={Du, Zhiying and Xing, Zhen},
  booktitle={ECCV Workshop},
  year={2024}
}

@article{li2022bevformer,
  title={BEVFormer: Learning Bird’s-Eye-View Representation from Multi-Camera Images via Spatiotemporal Transformers},
  author={Li, Zhiqi and others},
  journal={arXiv preprint arXiv:2203.17270},
  year={2022}
}

@article{li2024uniscene,
  title={UniScene: Unified Occupancy-centric Driving Scene Generation},
  author={Li, Bohan and others},
  journal={arXiv preprint arXiv:2412.05435},
  year={2024}
}

@article{liu2024timestep,
  title={Timestep Embedding Tells: It's Time to Cache for Video Diffusion Model},
  author={Liu, Feng and others},
  journal={arXiv preprint arXiv:2411.19108},
  year={2024}
}

@inproceedings{zhang2025sageattention,
      title={SageAttention: Accurate 8-Bit Attention for Plug-and-play Inference Acceleration}, 
      author={Zhang, Jintao and others},
      booktitle=ICLR,
      year={2025}
}

@misc{zhang2024sageattention2,
      title={SageAttention2: Efficient Attention with Thorough Outlier Smoothing and Per-thread INT4 Quantization}, 
      author={Jintao Zhang and others},
      year={2024},
      eprint={2411.10958},
      archivePrefix={arXiv},
      primaryClass={cs.LG},
      url={https://arxiv.org/abs/2411.10958}, 
}

@misc{oquab2023dinov2,
  title={DINOv2: Learning Robust Visual Features without Supervision},
  author={Oquab, Maxime and others},
  journal={arXiv:2304.07193},
  year={2023}
}

@inproceedings{dao2023flashattention2,
  title={Flash{A}ttention-2: Faster Attention with Better Parallelism and Work Partitioning},
  author={Dao, Tri},
  booktitle=ICLR,
  year={2024}
}

@article{wang2025zpressor,
  title={ZPressor: Bottleneck-Aware Compression for Scalable Feed-Forward 3DGS},
  author={Wang, Weijie and Chen, Donny Y and Zhang, Zeyu and Shi, Duochao and Liu, Akide and Zhuang, Bohan},
  journal={arXiv preprint arXiv:2505.23734},
  year={2025}
}

@inproceedings{zhang2026panflow,
  title={Panflow: Decoupled motion control for panoramic video generation},
  author={Zhang, Cheng and Liang, Hanwen and Chen, Donny Y and Wu, Qianyi and Plataniotis, Konstantinos N and Gambardella, Camilo Cruz and Cai, Jianfei},
  booktitle=AAAI,
  volume={40},
  number={15},
  pages={12385--12393},
  year={2026}
}

@article{shi2025pmloss,
  title={Revisiting Depth Representations for Feed-Forward 3D Gaussian Splatting},
  author={Shi, Duochao and Wang, Weijie and Chen, Donny Y and Zhang, Zeyu and Bian, Jia-Wang and Zhuang, Bohan and Shen, Chunhua},
  journal={arXiv preprint arXiv:2506.05327},
  year={2025}
}

@article{wang2025volsplat,
  title={Volsplat: Rethinking feed-forward 3d gaussian splatting with voxel-aligned prediction},
  author={Wang, Weijie and Chen, Yeqing and Zhang, Zeyu and Liu, Hengyu and Wang, Haoxiao and Feng, Zhiyuan and Qin, Wenkang and Chen, Feng and Zhu, Zheng and Chen, Donny Y and others},
  journal={arXiv preprint arXiv:2509.19297},
  year={2025}
}

@article{ni2025wonderturbo,
  title={Wonderturbo: Generating interactive 3d world in 0.72 seconds},
  author={Ni, Chaojun and others},
  journal={arXiv preprint arXiv:2504.02261},
  year={2025}
}

@inproceedings{yang2024unipad,
  title={Unipad: A universal pre-training paradigm for autonomous driving},
  author={Yang, Honghui and others},
  booktitle=CVPR,
  pages={15238--15250},
  year={2024}
}

@inproceedings{huang2024selfocc,
  title={Selfocc: Self-supervised vision-based 3d occupancy prediction},
  author={Huang, Yuanhui and others},
  booktitle=CVPR,
  pages={19946--19956},
  year={2024}
}

@article{wang2024distillnerf,
  title={Distillnerf: Perceiving 3d scenes from single-glance images by distilling neural fields and foundation model features},
  author={Wang, Letian and others},
  journal=NIPS,
  volume={37},
  pages={62334--62361},
  year={2024}
}

@article{xu2025resplat,
  title={ReSplat: Learning Recurrent Gaussian Splats},
  author={Xu, Haofei and others},
  journal={arXiv preprint arXiv:2510.08575},
  year={2025}
}

@article{duan2026liveworld,
  title={LiveWorld: Simulating Out-of-Sight Dynamics in Generative Video World Models},
  author={Duan, Zicheng and Xia, Jiatong and Zhang, Zeyu and Zhang, Wenbo and Zhou, Gengze and Gou, Chenhui and He, Yefei and Chen, Feng and Zhang, Xinyu and Liu, Lingqiao},
  journal={arXiv preprint arXiv:2603.07145},
  year={2026}
}

@article{wu2026light4d,
  title={Light4D: Training-Free Extreme Viewpoint 4D Video Relighting},
  author={Wu, Zhenghuang and Chen, Kang and Zhang, Zeyu and Tang, Hao},
  journal={arXiv preprint arXiv:2602.11769},
  year={2026}
}

@article{zhang2025egolcd,
  title={EgoLCD: Egocentric Video Generation with Long Context Diffusion},
  author={Zhang, Liuzhou and Ye, Jiarui and Wang, Yuanlei and Zhong, Ming and Cao, Mingju and Xia, Wanke and Zeng, Bowen and Zhang, Zeyu and Tang, Hao},
  journal={arXiv preprint arXiv:2512.04515},
  year={2025}
}

@article{zhang2025blockvid,
  title={BlockVid: Block Diffusion for High-Quality and Consistent Minute-Long Video Generation},
  author={Zhang, Zeyu and Chang, Shuning and He, Yuanyu and Han, Yizeng and Tang, Jiasheng and Wang, Fan and Zhuang, Bohan},
  journal={arXiv preprint arXiv:2511.22973},
  year={2025}
}

@article{luo2025univid,
  title={Univid: The open-source unified video model},
  author={Luo, Jiabin and Lin, Junhui and Zhang, Zeyu and Wu, Biao and Fang, Meng and Chen, Ling and Tang, Hao},
  journal={arXiv preprint arXiv:2509.24200},
  year={2025}
}

@article{liu2025fpsattention,
  title={Fpsattention: Training-aware fp8 and sparsity co-design for fast video diffusion},
  author={Liu, Akide and Zhang, Zeyu and Li, Zhexin and Bai, Xuehai and Han, Yizeng and Tang, Jiasheng and Xing, Yuanjie and Wu, Jichao and Yang, Mingyang and Chen, Weihua and others},
  journal={arXiv preprint arXiv:2506.04648},
  year={2025}
}

\end{document}